\documentclass[conference]{IEEEtran}
\IEEEoverridecommandlockouts
\usepackage{cite}
\usepackage{amsmath,amssymb,amsfonts}
\usepackage{algorithmic}
\usepackage{graphicx}
\usepackage{textcomp}
\usepackage{xcolor}
\usepackage{booktabs} 
\usepackage{multirow} 
\usepackage{siunitx} 
\usepackage[normalem]{ulem} 
\usepackage{tabularx} 
\def\BibTeX{{\rm B\kern-.05em{\sc i\kern-.025em b}\kern-.08em
    T\kern-.1667em\lower.7ex\hbox{E}\kern-.125emX}}
\begin{document}

\title{Polyp-VSSD: A Pre-trained Transformer with Local–Global Joint VSSD Decoding for Polyp Segmentation}

\author{%
\IEEEauthorblockN{%
Xiaobo Zhu\textsuperscript{1,2},
Shiwei Wu\textsuperscript{3},
Yu An\textsuperscript{4,5},
Yuechen Tao\textsuperscript{1,2},\\
He Sun\textsuperscript{4,5},
Song Zhang\textsuperscript{1,2},
and Zhenyu Liu\textsuperscript{1,2,*}
}

\IEEEauthorblockA{%
\textsuperscript{1}CAS Key Laboratory of Molecular Imaging and
Beijing Key Laboratory of Molecular Imaging,\\
Institute of Automation, Chinese Academy of Sciences,
Beijing 100190, China
}

\IEEEauthorblockA{%
\textsuperscript{2}School of Artificial Intelligence,
University of Chinese Academy of Sciences,
Beijing 100049, China
}

\IEEEauthorblockA{%
\textsuperscript{3}School of Computer Science and Engineering,
Nanjing University of Science and Technology,\\
Nanjing, Jiangsu 210094, China
}

\IEEEauthorblockA{%
\textsuperscript{4}School of Engineering Medicine,
Beihang University, Beijing 100191, China
}

\IEEEauthorblockA{%
\textsuperscript{5}Key Laboratory of Big Data-Based Precision Medicine
(Beihang University),\\
Ministry of Industry and Information Technology of China,
Beijing 100191, China
}

\IEEEauthorblockA{%
\textsuperscript{*}Corresponding author: Zhenyu Liu
}
}

\maketitle

\begin{abstract}
Early detection of colorectal polyps is essential for preventing colorectal cancer, making accurate polyp segmentation a key component in computer-aided diagnosis systems. However, existing endoscopic segmentation methods often exhibit limited representational capacity and insufficient integration of global and local contextual information. To address these limitations, we propose Polyp-VSSD, a novel local–global joint decoding framework built upon a pre-trained Transformer backbone for high-precision polyp segmentation. Polyp-VSSD leverages the strong feature expressiveness of large-scale visual pre-training and unifies global semantic modeling with local fine-grained reconstruction to enhance segmentation fidelity. Extensive experiments on multiple public endoscopic datasets demonstrate that our approach significantly outperforms state-of-the-art methods in both Dice and IoU. On the EndoScene dataset, Polyp-VSSD achieves mIoU and mDice scores of 0.903 and 0.946, respectively, highlighting its robustness and strong generalization capability. This work presents an effective and scalable solution for intelligent endoscopic image analysis. 
\end{abstract}


\section{Introduction}
\label{sec:intro}

Colorectal cancer (CRC) is one of the most common cancers globally, and its consistently high incidence and mortality rates present a significant threat to public health\cite{siegel2024cancer, siegel2023colorectal,10787150 }. Extensive clinical evidence shows that more than 90\% of CRC cases arise from the malignant transformation of adenomatous polyps, a process that typically progresses over several years\cite{vogelstein1988genetic}. Consequently, early screening via colonoscopy—enabling timely detection and removal of precancerous lesions, particularly colorectal polyps—remains an essential strategy for preventing CRC and reducing associated mortality\cite{winawer1993prevention,atkin2010once}.However, precisely detecting and segmenting polyps during colonoscopy remains challenging because of the wide variability in polyp morphology, size, and color, along with frequent imaging artifacts such as uneven illumination, specular highlights, motion blur, and residual bowel contents. Therefore, developing a highly accurate and robust automated polyp segmentation system is fundamental to building an effective Computer-Aided Diagnosis (CAD) framework, which can alleviate clinicians’ workload and enhance both diagnostic efficiency and accuracy.

In recent years, with the rapid advancement of deep learning—particularly convolutional neural networks (CNNs) and Transformer-based architectures—artificial intelligence (AI) has shown remarkable potential in medical image analysis\cite{wang2024comprehensive}. By learning complex visual representations, these models have been successfully applied to lesion detection, classification, and segmentation tasks. In polyp segmentation, deep learning–based approaches have substantially outperformed traditional image processing methods, leading to notable improvements in segmentation accuracy. For example, the U-Net architecture and its numerous variants\cite{ronneberger2015u}, as well as methods integrating attention mechanisms and cross-scale feature fusion\cite{zhou2018unet++}, can effectively extract polyp boundaries from complex endoscopic scenes through end-to-end feature learning. Despite these advancements, current models still fall short of fully addressing the intrinsic challenges of endoscopic imaging, such as the low contrast between polyps and surrounding mucosa, highly irregular polyp shapes, and domain shift across datasets collected under different clinical conditions.

To overcome the limitations of existing models in handling long-range dependencies and computational efficiency, State Space Models (SSMs) have attracted increasing attention due to their linear computational complexity. Mamba and its enhanced variant, Mamba2\cite{dao2024transformers}, achieve performance comparable to—or even exceeding—that of Transformers in sequence modeling tasks by incorporating selective state-space mechanisms and leveraging State Space Duality (SSD). The SSD framework uncovers an algorithmic-level duality between SSMs and structured masked attention, facilitating both computational acceleration and improved model capacity. Given that the original SSD/SSM formulation is inherently causal, we introduce Visual State Space Duality (VSSD)\cite{shi2025vssd} to extend these advantages to non-causal visual tasks. By adopting a non-causal SSD formulation and integrating multi-scan strategies, VSSD effectively adapts this efficient linear decoding paradigm to the visual domain. This establishes a new pathway for designing visual backbone networks that preserve strong performance while offering computational benefits comparable to or surpassing those of Transformer-based architectures. 

In the field of polyp segmentation, numerous deep learning-based methods have been proposed. PraNet\cite{fan2020pranet} addresses the challenges posed by polyp size variation and indistinct boundaries by introducing Parallel Partial Decoders (PPD) and Reverse Attention (RA) modules, which operate in a region–boundary coordination scheme to refine predictions and improve contour delineation. ColnNet\cite{jain2023coinnet} integrates convolution and deconvolution (involution) operations, further incorporating statistical feature attention and anomaly boundary approximation modules to enhance spatial feature discrimination and recursive boundary fusion, showing particular strength in detecting subtle or diminutive polyps. With the rise of large-scale pre-trained vision models, recent research trends have shifted toward leveraging foundation models for medical imaging. For example, ASPS\cite{li2024asps} exploits the strong generalization of the Segment Anything Model (SAM) by coupling trainable CNNs with frozen ViTs through a Cross-Branch Feature Augmentation (CFA) module, effectively narrowing the domain gap between natural and endoscopic images while recovering fine-grained local structures. Polyper\cite{shao2024polyper} further draws inspiration from clinical workflow to introduce a boundary-sensitive attention mechanism that uses interior features to enhance boundary discrimination. In parallel, CFA-Net\cite{zhou2023cross} develops a dual-stream framework consisting of a boundary prediction network and Cross-Level Feature Fusion (CFF) modules, enabling the joint aggregation of multi-scale and cross-level semantics to better handle polyp size variability. 

However, although these models have substantially advanced polyp segmentation, they continue to face several fundamental limitations\cite{qayoom2025polyp}. First, the encoder is often trained from scratch or subjected only to limited pre-training on small publicly available polyp datasets (e.g., Kvasir-SEG, CVC-ClinicDB). Such restricted training leads to insufficient generalization ability, preventing the encoder from learning robust representations that can adapt to complex endoscopic environments and diverse polyp morphologies—particularly when encountering unseen or rare polyp appearances. Second, existing decoder designs often fail to adequately exploit global semantic information and local structural details in a unified manner. Many rely on relatively simple feature fusion strategies and thus struggle to simultaneously leverage long-range global context (e.g., overall polyp location and shape) for linear decoding while reconstructing fine-grained, high-resolution local details (e.g., precise polyp boundaries). This imbalance ultimately undermines segmentation accuracy and reduces overall robustness.

To address these challenges, we propose Polyp-VSSD, a polyp segmentation framework that integrates a large-scale pre-trained Transformer backbone with a Local–Global Joint VSSD Decoding (LG-VSSD) mechanism. First, we pre-train the Transformer encoder using the DINOv2\cite{oquab2023dinov2} self-supervised learning paradigm on a large-scale private endoscopic video dataset exceeding 20,000 minutes. This enables the encoder to acquire robust and highly generalizable representations tailored to diverse endoscopic morphologies. Second, we carefully design the LG-VSSD decoder, which incorporates a lightweight CNN branch to recover fine-grained local structures and a VSSD-based branch to perform efficient linear decoding of long-range global features. By jointly leveraging local detail reconstruction and global semantic modeling, the proposed decoder yields segmentation outputs that preserve precise boundary information while maintaining rich contextual understanding. Moreover, we conduct extensive evaluations across five public datasets—CVC-ClinicDB, Kvasir-SEG, CVC-ColonDB, ETIS-LaribPolypDB, and EndoScene. The experimental results show that Polyp-VSSD significantly surpasses existing state-of-the-art methods across key metrics such as Dice and IoU. In particular, on the challenging EndoScene dataset, Polyp-VSSD attains an mIoU of 0.903 and an mDice of 0.946, demonstrating outstanding robustness and strong generalization capability. Our contributions are summarized as follows:
\begin{itemize}
    \item Through self-supervised pre-training on a large-scale private endoscopy dataset, a novel Transformer encoder is obtained, exhibiting strong generalization and significantly enhanced polyp feature representation.
    \item An innovatively designed local–global joint VSSD decoder (LG-VSSD) integrates the detail-preserving strengths of the CNN branch with the global linear decoding efficiency of the VSSD branch, enabling high-precision and high-performance polyp segmentation.
    \item Extensive and systematic evaluations over five mainstream public polyp segmentation benchmarks demonstrate that Polyp-VSSD delivers state-of-the-art results, showing consistent superiority across all datasets.
\end{itemize}

\section{METHODS}
\label{sec:methods}

\subsection{Endoscopic Foundation Model Pre-training}
\begin{figure*}[htbp]
\centering
\includegraphics[width=0.95\linewidth]{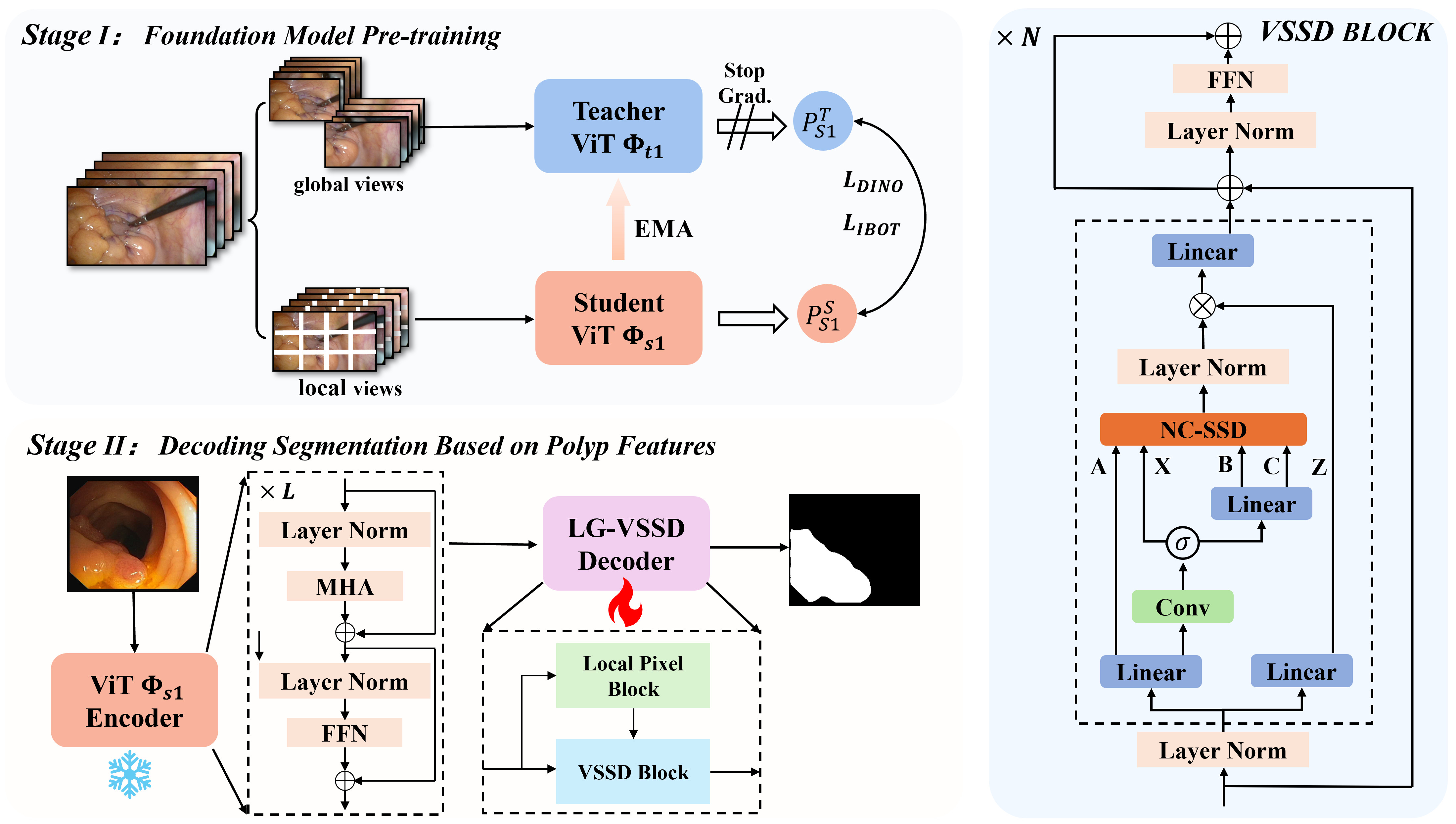}
\caption{Overview of the Polyp-VSSD framework. Stage I: Pre-train an endoscopic foundation encoder to learn modality-specific visual priors; stage II: Decode encoder features via the LG-VSSD module to jointly enhance global semantics and local boundary representation for final polyp segmentation.}
\label{fig: framework}
\end{figure*}

To address the limited generalization capacity of existing polyp segmentation encoders, this work adopts a self-supervised pre-training paradigm on large-scale private endoscopic video data to construct a robust endoscopic foundation model, as shown in Fig.\ref{fig: framework}. Conventional approaches, typically trained on small and heterogeneous public datasets, often struggle to capture the diverse and complex visual attributes inherent to endoscopy—such as variability in polyp morphology, illumination shifts, specular reflections, mucosal texture variation, and motion-induced blur.

To mitigate these limitations, the DINOv2 (Self-Distillation with Vision Transformers) framework is employed for large-scale pre-training using over 20,000 minutes of raw endoscopic video. To reduce temporal redundancy, video streams are uniformly sampled at two frames per second, yielding approximately 1.4 million unlabeled frames for representation learning. DINOv2 enables dense feature learning without manual annotations through a knowledge-distillation mechanism between teacher and student networks. In implementation, a standard Vision Transformer (ViT) serves as the encoder backbone and is optimized via the DINOv2 objective, which enforces consistency between student predictions and the teacher's smoothed output distribution.

Following extensive self-supervised pre-training, the resulting Transformer encoder internalizes endoscopy-specific visual priors and contextual dependencies, thereby enhancing representation quality with respect to intrinsic polyp morphology, boundary structure, texture transitions, and background suppression. This produces a more discriminative and robust feature space for downstream segmentation. The pre-trained ViT encoder is subsequently integrated into the Polyp-VSSD framework as the feature extraction backbone.
All manuscripts must be in English.

\subsection{SSD Preliminaries}
The classical state-space model (SSM) describes continuous-time system dynamics by mapping an input signal \(x(t) \in \mathbb{R}\) to a latent state \(h(t) \in \mathbb{R}^N\), which then produces the output \(y(t) \in \mathbb{R}\). The model is formulated as
\begin{equation}
\frac{d h(t)}{dt} = \overset{\circ}{\mathbf{A}}\, h(t) + \overset{\circ}{\mathbf{B}}\, x(t), 
\qquad
y(t) = \mathbf{C}\, h(t),
\end{equation}
where \( \overset{\circ}{\mathbf{A}} \in \mathbb{R}^{N \times N} \), 
\( \overset{\circ}{\mathbf{B}} \in \mathbb{R}^{N \times 1} \), and 
\( \mathbf{C} \in \mathbb{R}^{1 \times N} \) denote the system parameters\cite{dao2024transformers}.
To incorporate continuous SSMs into deep learning architectures, discretization is required. This is achieved by introducing a time-step parameter \( \Delta \in \mathbb{R} \) and applying 
zero-order hold (ZOH), which converts the continuous matrices \( \overset{\circ}{\mathbf{A}} \) and \(\overset{\circ}{\mathbf{B}}\) into their discrete counterparts \( {\mathbf{A}} \) and \( {\mathbf{B}}\)\cite{shi2025vssd}.

Mamba2\cite{dao2024transformers} introduced the State Space Dual (SSD), reducing the state transition matrix \({\mathbf{A}}\) to a scalar, which forms a special case of the Selective SSM that supports both linear and quadratic formulations. Without loss of generality, the matrix form of the SSM is expressed as:
\begin{equation}
\begin{aligned}
& y(t)=\sum_{i=1}^t \mathbf{C}_t^T \mathbf{A}_{t: i+1} \mathbf{B}_i x(i), \text { where } \mathbf{A}_{t: i}=\prod_{i=2}^t \mathbf{A}_i, \\
& y=\operatorname{SSM}(\mathbf{A}, \mathbf{B}, \mathbf{C})(x)=\mathbf{F} x, \text { where } \mathbf{F}_{j i}=\mathbf{C}_j^T \mathbf{A}_{j: i} \mathbf{B}_i .
\end{aligned}
\label{eq:ssm_quadratic}
\end{equation}
where $\mathbf{A}$ is simplified to a scalar, the quadratic form of (\ref{eq:ssm_quadratic}) can be rewritten as:
\begin{equation}
\begin{aligned}
&y=\mathbf{F} x=\mathbf{M} \cdot\left(\mathbf{C}^T \mathbf{B}\right) x, \\
&\text { where } \mathbf{M}_{i j}= \begin{cases}A_{i+1} \times \cdots \times A_j & i>j \\ 1 & i=j \\ 0 & i<j\end{cases}
\end{aligned}
\end{equation}
while its linear form is denoted as:
\begin{equation}
h(t)=A_t h(t-1)+\mathbf{B}_t x(t), y(t)=\mathbf{C}_t h(t) .
\end{equation}

The inherent causality of traditional SSMs restricts each token to interacting only with past inputs, resulting in a unidirectional information flow. Such a constraint is unsuitable for visual tasks, where spatial interactions are inherently non-causal. In addition, reshaping a 2D feature map into a 1D sequence disrupts its spatial topology—tokens that are neighbors in 2D may become distant in the flattened sequence—ultimately weakening the model’s ability to capture local spatial relationships and degrading segmentation performance.
\begin{table*}[htbp] 
\centering
\caption{Comparison with existing polyp segmentation methods on five public datasets.
Bold denotes the best result, and underlined denotes the second-best result.}
\label{tab:comparison_results}
\begin{tabularx}{0.9\linewidth}{c | c | *{4}{S[table-format=1.3] S[table-format=1.3] |} S[table-format=1.3] S[table-format=1.3]}
\toprule
\multirow{2}{*}{\textbf{Methods}} & \multirow{2}{*}{\textbf{Venue}} & \multicolumn{2}{c}{\textbf{CVC-ClinicDB}} & \multicolumn{2}{c}{\textbf{Kvasir-Seg}} & \multicolumn{2}{c}{\textbf{CVC-ColonDB}} & \multicolumn{2}{c}{\textbf{ETIS}} & \multicolumn{2}{c}{\textbf{EndoScene}} \\
\cmidrule(lr){3-4} \cmidrule(lr){5-6} \cmidrule(lr){7-8} \cmidrule(lr){9-10} \cmidrule(lr){11-12}
& & {mIoU} & {mDice} & {mIoU} & {mDice} & {mIoU} & {mDice} & {mIoU} & {mDice} & {mIoU} & {mDice} \\
\midrule
PraNet & MICCAI'20 & 0.849 & 0.899 & 0.840 & 0.898 & 0.640 & 0.709 & 0.567 & 0.628 & 0.797 & 0.871 \\
SSFormer & MICCAI'22 & 0.855 & 0.906 & 0.864 & 0.917 & 0.721 & 0.802 & 0.720 & 0.796 & 0.827 & 0.895 \\
CFA-Net & PR'23 & 0.883 & 0.933 & 0.861 & 0.915 & 0.665 & 0.743 & 0.655 & 0.732 & 0.827 & 0.893 \\
CoInNet & TMI'23 & 0.887 & 0.930 & 0.872 & 0.926 & 0.729 & 0.797 & 0.690 & 0.759 & 0.863 & 0.909 \\
ASPS & MICCAI'24 & 0.891 & 0.942 & 0.849 & 0.914 & 0.680 & 0.782 & 0.758 & 0.854 & 0.819 & 0.900 \\
LSSNet & MICCAI'24 & \underline{0.900} & \underline{0.945} & 0.878 & 0.926 & 0.741 & 0.820 & 0.775 & 0.846 & 0.822 & 0.894 \\
Polyper & AAAI'24 & 0.899 & 0.944 & 0.856 & 0.920 & 0.745 & \underline{0.837} & \underline{0.785} & \underline{0.865} & 0.867 & \underline{0.924} \\
Polyp-Mamba & IF'24 & 0.896 & 0.941 & 0.867 & 0.919 & 0.713 & 0.791 & 0.756 & 0.668 & 0.840 & 0.906 \\
UM-Net & MIA'25 & 0.888 & 0.913 & \underline{0.879} & \underline{0.930} & \underline{0.761} & 0.828 & 0.645 & 0.785 & \underline{0.893} & 0.903 \\
ODR & TM'25 & 0.894 & 0.940 & 0.868 & 0.920 & 0.732 & 0.811 & 0.721 & 0.795 & 0.842 & 0.907 \\
MSBP-Net & PR'25 & 0.892 & 0.940 & 0.868 & 0.919 & 0.731 & 0.810 & 0.718 & 0.795 & 0.828 & 0.897 \\
\midrule
SurgicalSAM & AAAI'24 & 0.505 & 0.644 & 0.597 & 0.740 & 0.330 & 0.460 & 0.238 & 0.342 & 0.472 & 0.623 \\
MedSAM & Nat. Commun'24 & 0.803 & 0.867 & 0.795 & 0.862 & 0.651 & 0.734 & 0.604 & 0.687 & 0.798 & 0.870 \\
WeakPolyp-SAM & KBS'25 & 0.805 & 0.882 & 0.797 & 0.869 & 0.621 & 0.700 & 0.603 & 0.691 & 0.813 & 0.886 \\
OP-SAM & ICCV'25 & 0.678 & 0.756 & 0.769 & 0.845 & 0.608 & 0.681 & 0.687 & 0.755 & 0.687 & 0.779 \\
\midrule
\textbf{Polyp-VSSD(ours)} & {--} & \textbf{0.923} & \textbf{0.959} & \textbf{0.881} & \textbf{0.935} & \textbf{0.764} & \textbf{0.852} & \textbf{0.830} & \textbf{0.899} & \textbf{0.903} & \textbf{0.946} \\
\bottomrule
\end{tabularx}
\end{table*} 

\subsection{Local-Global Visual Spatial State Duality Module}
To extend SSD to non-causal visual modeling, the VSSD (Visual State Space Duality)\cite{shi2025vssd} architecture is adopted. As a non-causal variant of SSD, VSSD removes the absolute interaction magnitude between hidden states and input tokens while retaining their relative weighting, thereby decoupling token dependencies and enabling non-causal feature aggregation. To capture full 2D spatial context, each VSSD block applies multi-scan strategies, performing bidirectional or multi-directional scans over flattened image sequences and executing the SSD mechanism within each scan path. The scan outputs are then fused to ensure that every spatial token integrates holistic, non-causal context from the entire image. With this mechanism, the decoder achieves efficient long-range global feature fusion with linear complexity, substantially enhancing polyp localization and semantic perception. The theoretical formulation of VSSD is provided in the supplementary material.

Additionally, a CNN-based pyramid structure is integrated as a local pixel block (LPB) to further enhance decoding capability, as shown in Fig.\ref{fig: framework}. Leveraging CNN’s strength in local feature perception, multi-scale encoded representations are progressively upsampled and injected into the VSSD decoder as complementary local cues. This joint decoding mechanism enriches fine-grained spatial details and thereby improves segmentation precision. The process is formulated as:
\begin{equation}
\begin{aligned}
F_{1} &= LPB_{0}(F_{\text{encoder}}) + VSSD(F_{\text{encoder}}), \quad i=0, \\
F_{i} &= VSSD\!\left(F_{1} + LPB_{i}(F_{\text{encoder}})\right), \quad i \ge 1
\end{aligned}   
\end{equation}
where i denotes the index of the decoder layer, while LPB and VSSD correspond to the Local Pixel Block and Visual Spatial State Duality module, respectively. By jointly leveraging these two components, the proposed Local–Global Joint VSSD (LG-VSSD) decoder strengthens the extraction and reconstruction of fine-grained polyp morphology. The LPB efficiently recovers local structural details, whereas the VSSD performs linear-complexity global scanning to aggregate long-range contextual information. Through this coupling of local enhancement and global perception, the decoder achieves more accurate feature restoration and ultimately yields superior segmentation performance.

\section{EXPERIMENT}
\label{sec:exper}
\begin{table*}[htbp] 
\centering
\caption{Ablation study on validation of the effectiveness of the VSSD module and Local Pixel Block (LPB).}
\label{tab:ablation_results}
\begin{tabularx}{0.9\linewidth}{c | c | *{4}{S S |} S S}
\toprule
\multicolumn{2}{c|}{\textbf{Modules}} & \multicolumn{2}{c|}{\textbf{CVC-ClinicDB}} & \multicolumn{2}{c|}{\textbf{Kvasir-Seg}} & \multicolumn{2}{c|}{\textbf{CVC-ColonDB}} & \multicolumn{2}{c|}{\textbf{ETIS}} & \multicolumn{2}{c}{\textbf{EndoScene}} \\
\cmidrule(lr){1-2} \cmidrule(lr){3-4} \cmidrule(lr){5-6} \cmidrule(lr){7-8} \cmidrule(lr){9-10} \cmidrule(lr){11-12}
\textbf{VSSD} & \textbf{LPB} & {\quad mIoU} & {\quad mDice} & {\quad mIoU} & {\quad mDice} & {\quad mIoU} & {\quad mDice} & {\quad mIoU} & {\quad mDice} & {\quad mIoU} & {\quad mDice} \\
\midrule
-- & -- & 0.881 & 0.934 & 0.842 & 0.911 & 0.739 & 0.831 & 0.796 & 0.875 & 0.877 & 0.931 \\
$\checkmark$ & -- & 0.904 & 0.947 & 0.868 & 0.927 & 0.786 & 0.841 & 0.814 & 0.888 & 0.893 & 0.941 \\
-- & $\checkmark$ & 0.895 & 0.942 & 0.858 & 0.921 & 0.750 & 0.833 & 0.809 & 0.884 & 0.889 & 0.938 \\
$\checkmark$ & $\checkmark$ &\textbf{\quad0.923} & \textbf{\quad0.959} & \textbf{\quad0.881} & \textbf{\quad0.935} & \textbf{\quad0.764} & \textbf{\quad0.852} & \textbf{\quad0.830} & \textbf{\quad0.899} & \textbf{\quad0.903} & \textbf{\quad0.946} \\
\bottomrule
\end{tabularx}
\end{table*}

\subsection{Datasets and Experimental Settings}
The experimental data comprise two primary categories: large-scale endoscopic video data for encoder pretraining and public polyp segmentation benchmarks for downstream evaluation. To build a high-quality, modality-specific pretraining set, more than 20,000 minutes of private endoscopy videos underwent rigorous filtering. Non-clinical scenes, blurry or low-quality frames, and visually redundant clips were removed to maximize the relevance of supervision-free visual priors. To further reduce redundancy, frames with minimal temporal variations were discarded before applying a uniform sampling strategy at 2 FPS, resulting in 1,451,670 cleaned endoscopic images, all used for DINOv2-based self-supervised pretraining.

Following pretraining, polyp segmentation was trained and evaluated following benchmark protocols consistent with PraNet\cite{fan2020pranet} for fair comparison. A total of 1,450 labeled images were used for training—900 from Kvasir-Seg and 550 from CVC-ClinicDB. Model generalization performance was assessed on five challenging public datasets comprising 698 images across diverse acquisition settings and polyp characteristics: CVC-ClinicDB, Kvasir-Seg, CVC-ColonDB, EndoScene, and ETIS-LaribPolypDB. 

The full experimental pipeline consists of two stages—encoder pretraining and polyp segmentation training—outlined in Figure 1. Base pretraining was performed on four NVIDIA A100 GPUs for 100 epochs with a learning rate of $1\times10^{-6 } $ and weight decay of 0.04, following the official DINOv2 multi-view augmentation scheme (two global + eight local crops). The Polyp-VSSD model was then trained for 100 epochs on a single A100 GPU, with the encoder frozen and only the decoder optimized using a learning rate of $2\times10^{-4 } $ and weight decay of 0.001. Both the LPB and VSSD components of the decoder employed a depth of three layers.

\subsection{Comparative Study with Multiple Models}
To validate the effectiveness of the proposed method, we conduct extensive comparisons with a broad range of state-of-the-art (SOTA) polyp segmentation approaches, including CNN-based, Transformer-based, and Mamba-based segmentation frameworks—such as PraNet\cite{fan2020pranet}, SSFormer\cite{wang2022stepwise}, CFA-Net\cite{zhou2023cross}, CoInNet\cite{jain2023coinnet}, ASPS\cite{li2024asps}, LSSNet\cite{wang2024lssnet}, Polyper\cite{shao2024polyper}, Polyp-Mamba\cite{zhu2025polyp}, UM-Net\cite{du2025net}, ODR\cite{li2025oriented}, and MSBP-Net\cite{pan2026msbp}—as well as SAM-based medical segmentation variants including SurgicalSAM\cite{yue2024surgicalsam}, MedSAM\cite{ma2024segment}, WeakPolyp-SAM\cite{zhao2025weakpolyp}, and OP-SAM\cite{mao2025one}. Quantitative results are summarized in Table \ref{tab:comparison_results}.

Our Polyp-VSSD model, which couples a pre-trained base encoder with LG-VSSD decoding, consistently achieves the best performance across all five benchmark datasets. Notably, it yields particularly large gains on ETIS and EndoScene, reaching mDice scores of 0.899 and 0.946, respectively. These results suggest that polyp-specific architectures are generally more effective than SAM-derived segmentation models—likely due to their stronger prior knowledge of endoscopic imaging and more informative prompt representations. This highlights the importance of encoders with modality-aware feature representations for downstream segmentation decoders.

\begin{figure}[htbp]
\centering
\includegraphics[width=\linewidth]{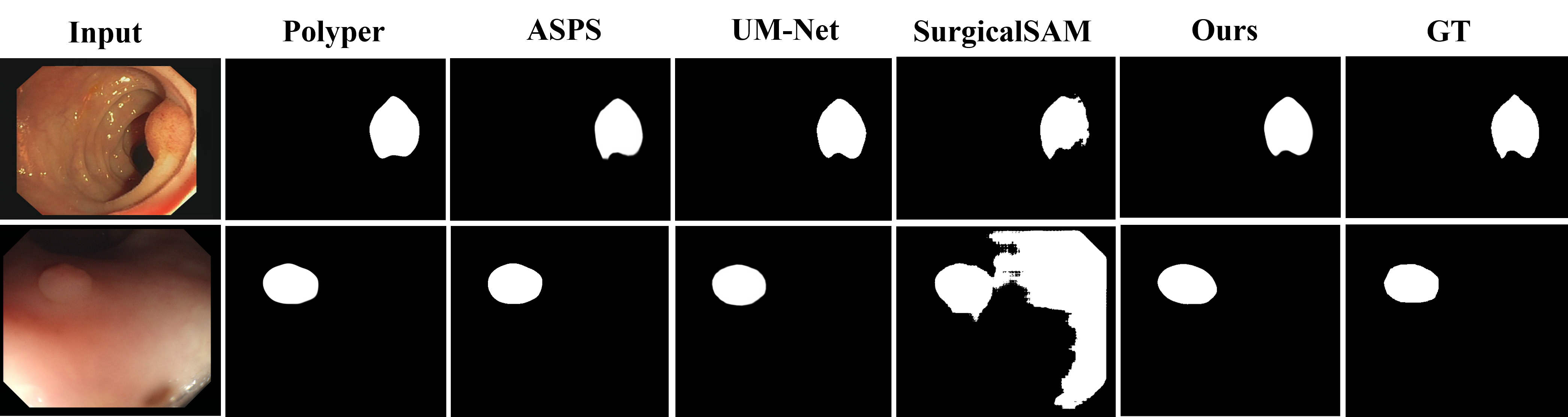}
\caption{Visual comparison of segmentation results across different methods}
\label{fig: Seg_results}
\end{figure}

Furthermore, while most existing methods excel selectively on certain datasets, they often struggle to maintain robustness across diverse clinical domains. For example, LSSNet performs strongly on ClinicDB (mIoU 0.900, mDice 0.945) yet exhibits reduced performance elsewhere. In contrast, Polyp-VSSD achieves the highest scores on every dataset, demonstrating superior generalization, more precise polyp boundary delineation, and enhanced structural perception, as visually confirmed in Fig.\ref{fig: Seg_results}.
\subsection{Ablation Study}
To assess the contribution of each core component in the Polyp-VSSD framework, we conducted a comprehensive series of ablation studies, with the results summarized in Table\ref{tab:ablation_results}.
The baseline configuration directly links the linear projection head of the pre-trained ViT encoder to the segmentation output. This setting yields relatively low mean Dice scores across all datasets, suggesting that limited decoding capability constrains the model from adequately restoring fine-grained polyp features.

Based on this baseline, we gradually incorporated the proposed LG-VSSD decoding modules. Introducing the VSSD global decoding path to replace the conventional CNN-based upsampling leads to consistent and notable performance gains. This confirms that VSSD, enabled by its non-causal duality formulation, effectively models long-range global dependencies with linear computational complexity, thereby reinforcing semantic comprehension of polyp morphology and spatial distribution.

Subsequently, we integrate the LPB module, which is designed as a pyramid-structured convolutional block for local feature refinement. This module directly enhances boundary-aware representation and improves reconstruction of subtle structures along polyp contours. The combined effect substantially boosts edge-level segmentation accuracy while maintaining strong global awareness.

With both the global VSSD path and the local LPB pathway active, the final Polyp-VSSD architecture establishes a Local–Global Joint VSSD decoding mechanism that achieves state-of-the-art performance across all evaluation metrics. These results demonstrate that the LG-VSSD design effectively balances global semantic reasoning with high-fidelity local detail restoration: the global branch excels at capturing broad contextual relationships, while the local branch focuses on precise structure delineation. Together, they enable highly reliable segmentation of polyps with diverse morphology and complex boundaries.

\section{CONCLUSION}
This study proposes an innovative decoupled polyp segmentation framework named Polyp-VSSD, aiming to balance segmentation accuracy and computational efficiency while enhancing the model's generalization capability in complex endoscopic environments. We first leveraged a large-scale private endoscopic video dataset to construct a robust endoscopic foundation model as an encoder through DINOv2 self-supervised pretraining, embedding modality-specific visual prior knowledge. Subsequently, we designed the LG-VSSD (Local-Global Joint VSSD Decoding) mechanism, which innovatively combines efficient global context modeling based on visual spatial state duality (VSSD) with a lightweight convolutional neural network for local refinement. Ablation studies demonstrate substantial synergistic gains from both the foundation model pretraining and the LG-VSSD mechanism on final performance. Quantitative comparisons reveal that Polyp-VSSD achieves SOTA performance across five challenging public polyp datasets, particularly outperforming existing methods on the morphologically irregular ETIS-LaribPolypDB and the complex background EndoScene datasets. In summary, Polyp-VSSD offers a novel and promising solution for efficiently and accurately processing endoscopic image segmentation tasks. In future research, we will explore further optimizations of the VSSD module to support real-time video processing and strive to extend this endoscopic base model to broader downstream applications such as polyp detection, classification, and multimodal fusion. This aims to provide more comprehensive and efficient support for clinical endoscopy-assisted diagnosis.

\bibliographystyle{IEEEbib}
\bibliography{icme2026references}

\end{document}